\documentclass{article}

\usepackage{arxiv}

\usepackage[utf8]{inputenc} 
\usepackage[T1]{fontenc}    
\usepackage{hyperref}       
\usepackage{url}            
\usepackage{booktabs}       
\usepackage{amsfonts}       
\usepackage{nicefrac}       
\usepackage{microtype}      
\usepackage{lipsum}
\usepackage{mathtools}
\usepackage[english]{babel}
\usepackage{blindtext,tikz}

\title{Comparisons of Graph Neural Networks on Cancer Classification Leveraging a Joint of Phenotypic and Genetic Features}

\author{
  David Oniani \\
  Mayo Clinic\\
  Rochester, MN 55902 \\
  \texttt{oniada01@luther.edu} \\
  \And
 Chen Wang \\
  Mayo Clinic\\
  Rochester, MN 55902 \\
  \texttt{wang.chen@mayo.edu} \\
    \And
 Yiqing Zhao \\
  Mayo Clinic\\
  Rochester, MN 55902 \\
  \texttt{zhao.yiqing@mayo.edu} \\
         \And
 Andrew Wen \\
  Mayo Clinic\\
  Rochester, MN 55902 \\
  \texttt{wen.andrew@mayo.edu} \\
     \And
 Hongfang Liu \\
  Mayo Clinic\\
  Rochester, MN 55902 \\
  \texttt{liu.hongfang@mayo.edu} \\
  
     \And
 Feichen Shen \\
  Mayo Clinic\\
  Rochester, MN 55902 \\
  \texttt{shen.feichen@mayo.edu}
  }
  
\usetikzlibrary{calc}
\begin{document}


\maketitle

\begin{abstract}
Cancer is responsible for millions of deaths worldwide every year. Although significant progress has been
achieved in cancer medicine, many issues remain to be addressed for improving cancer therapy. Appropriate cancer patient stratification is the prerequisite for selecting appropriate treatment plan, as cancer patients are of known heterogeneous genetic make-ups and phenotypic differences. In this study, built upon deep phenotypic characterizations extractable from Mayo Clinic electronic health records (EHRs) and genetic test reports for a collection of cancer patients, we evaluated various graph neural networks (GNNs) leveraging a joint of phenotypic and genetic features for cancer type classification. Models were applied and fine-tuned on the Mayo Clinic cancer disease dataset. The assessment was done through the reported accuracy, precision, recall, and F1 values as well as through F1 scores based on the disease class. Per our evaluation results, GNNs on average
outperformed the baseline models with mean
statistics always being higher that those
of the baseline models (0.849 vs 0.772 for
accuracy, 0.858 vs 0.794 for precision,
0.843 vs 0.759 for recall, and 0.843 vs
0.855 for F1 score). Among GNNs, ChebNet,
GraphSAGE, and TAGCN showed the best
performance, while GAT showed the worst. We applied and compared eight GNN models
including AGNN, ChebNet, GAT, GCN, GIN,
GraphSAGE, SGC, and TAGCN on the Mayo Clinic
cancer disease dataset and assessed their
performance as well as compared them with each
other and with more conventional machine
learning models such as decision tree,
gradient boosting, multi-layer perceptron,
naive bayes, and random forest which we
used as the baselines.
\end{abstract}


\section{Introduction}
Cancer is a type of disease in which abnormal cells divide without control and can invade nearby tissues and other parts of the body via the blood and lymph systems \cite{yuan2016deepgene,feuerstein2007defining}. Nowadays, cancer is responsible for millions of deaths worldwide every year. Although significant progress has been achieved in cancer medicine, many issues remain to be addressed for improving cancer therapy. The goal of Precision Oncology is to enable oncologist practitioners to make better clinical decisions by incorporating individual cancer patients’ genetic information and clinical characteristics \cite{schwartzberg2017precision}. Precision Oncology is expected to improve selection of targeted therapies, tailored to individual patients and ultimately improve cancer patients’ outcomes. However, patient heterogeneity needs to be considered while providing cancer treatments. It is important to take patient subgroups into consideration in which the mechanism of diseases with the same group are more likely to be homogeneous \cite{on2014improving}. Therefore, appropriate cancer patient subgrouping is the prerequisite for selecting appropriate treatment plan, as cancer patients are of known heterogeneous genetic make-ups and phenotypic differences. 

Advances in cancer care have demonstrated the benefits of Individualized Medicine research, and the premise of Precision Oncology is to refer similar cancer outcomes to similar patients with similar features \cite{parimbelli2018patient}. Phenotypic characterizations are one of the key features to group patients by analyzing their clinical observations \cite{seligson2020recommendations}. Taking breast cancer as an example, specific sub-phenotypes have been leveraged to make decision making for therapies selection in practice \cite{lehmann2011identification}. Moreover, phenotypic heterogeneity amongs 7 subtypes of triple-negative breast cancer could be observed and analyzed \cite{masuda2013differential}. With the advance of machine learning techniques, computational deep phenotyping is widely used and described as the process to analyze phenotypic abnormalities precisely and comprehensively, in order to receive a better understanding of the natural history of a disease and its genotype-phenotype associations. For example, the DeepPhe information model is able to extract cancer-related phenotypes as categorical features by mining clinical narratives or structured datasets \cite{savova2017deepphe}. Beeghly-Fadiel et al. have extended DeepPhe to ovarian cancer \cite{beeghly2019deep}. Besides categorical phenotypic features, embeddings representations of phenotypes are also significant to deep phenotyping. HPO2Vec+ provides a way to learn node embeddings for the Human Phenotype Ontology (HPO) and be able to represent patients with vectorized representations for individual-level phenotypic similarity measurement \cite{shen2019hpo2vec+,shen2018constructing}. Maldonado et al. presented an approach that uses Generative Adversarial Networks (GANs) \cite{goodfellow2017generative} for learning UMLS embeddings and demonstrated the use case on clinical prediction tasks \cite{maldonado2019adversarial}.

In addition to phenotypic characterizations, features contained in genetic test reports provide crucial evidence for the decision making in practice. Genetic tests for targeted cancer therapy detect mutations in the DNA of cancer cells. Being aware of whether the cancer has a particular mutation can help guide the type of treatment. \cite{sawyers2004targeted}. Specifically, pathogenic variants demonstrate the genetic alteration that increases an individual's susceptibility or predisposition to a certain cancer type, and variant of uncertain significance (VUS) indicates a genetic variant that has been identified but without knowing too much details regarding significance to the function of an organism. There exist some studies that use machine learning to support variant interpretation in a clinical setting. For example, the LEAP model classifies variants based on features collected from functional predictions, splice predictions, clinical observation data and so on \cite{lai2020leap}. Some guidelines to define variant pathogenicity for somatic and VUS classification were established \cite{federici2020variants,richards2015standards,rehm2013acmg}. Mut2Vec provides a way to represent cancerous mutations with distributed embeddings, in order to support cancer classification and drug sensitivity prediction \cite{kim2018mut2vec}.

Graph Neural Networks (GNNs) \cite{zhou2018graph} represent a series of deep learning methods that applied on graph domain. GNNs has been largely adopted for graph data characterization and analysis due to the promising performance. In this study, we constructed a graph leveraging phenotypic characterization and genetic features collected from patients and then compared performance on cancer type classification using various exisiting GNNs. Specifically, we compared performance amongst 8 GNNs, including Attentive Graph Neural Network (AGNN), Chebyshev Graph Neural Network (ChebNet), Graph Convolutional Network (GCN), Graph Attention Network (GAT), Graph Isomorphism Network (GIN), Graph SAmple and aggreGatE (GraphSAGE), Simplifying Graph Convolutional Network (SGC), and Topology Adaptive Graph Convolutional Network (TAGCN).

\section{Materials}
\label{sec:headings}
We collected a cancer patient cohort (n=794) with accessible Mayo Clinic EHRs and foundation genetic test reports. Seven types of cancer are included in this cohort: lung cancer (n=223), prostate cancer (n=66), breast cancer (n=53), ovarian cancer (n=91), pancreas cancer (n=104), colon/rectum cancer (n=140), and liver cancer (n=107).

\subsection{Extract phenotypic features from the Mayo Clinic EHRs}
Given patient IDs retrieved from the cohort, we first applied an existing Mayo Clinic NLP-as-a-service implementation \cite{wen2019desiderata} to extract all the diagnostic-related terms from clinical notes and radiology report, normalized by the Unified Medical Language System (UMLS) \cite{bodenreider2004unified}. We then used a previously developed Human Phenotype Ontology (HPO)-based annotation pipeline \cite{shen2017phenotypic} to further map the UMLS terms to more specific HPO normalized phenotypic terms \cite{robinson2008human}. Specifically, for each cancer patient, we collected their phentoypic information within the last twelve months of the first mention of cancer diagnoses.

\subsection{Collect genetic feature from the FoundationOne CDx genetic test reports}
FoundationOne CDx is a next-generation sequencing (NGS) based assay that identifies genomic findings within hundreds of cancer-related genes. The foundationOne genetic test report is designed to provide physicians with clinically actionable information. The report for each patient is represented in the PDF format. We first used a python library named pdfminer to parse all the PDF report into text format, for the convenience of data processing. We then parsed out genes alteration and VUS sections for each patient. In particular, we treated mutations mined from the genes alteration section as pathogenic mutation, and we named mutations mined from the VUS section as VUS mutation. We used multi-hot encoding to represent genetic features for each patient, where 1 indicates the existence of pathogenic mutation or VUS, and 0 otherwise.

\section{Methods}
We applied eight different machine learning models and assessed their
performance on Mayo Clinic cancer disease dataset. The main goal was to
conduct cancer disease classification tasks and evaluate the performance of
the models as well as to compare the models with each other.

The models included:

\begin{itemize}
    \item Attentive Graph Neural Network (AGNN)~\cite{agnn}
    \item Chebyshev Graph Neural Network (ChebNet)~\cite{chebnet}
    \item Graph Convolutional Network (GCN)~\cite{gcn}
    \item Graph Attention Network (GAT)~\cite{gat}
    \item Graph Isomorphism Network (GIN)~\cite{gin}
    \item Graph SAmple and aggreGatE (GraphSAGE)~\cite{graphsage}
    \item Simplifying Graph Convolutional Network (SGC)~\cite{sgc}
    \item Topology Adaptive Graph Convolutional Network (TAGCN)~\cite{tagcn}
\end{itemize}

We also compared them to the baseline models such as decision tree,
gradient boosting, multi-layer perceptron, naive bayes, and random forest.
All of the models were fine-tuned. We present the results in the results section.

\subsection{Data Preprocessing and Graph Generation}
For phenotypes, we first removed some general phenotypic terms mined from EHRs, including cyst, pain, carcinoma, neoplasm, symptoms, disease, minor (disease), and sarcoma - cateogry (morphologic abnormality). We then filtered out phenotypic terms with low frequency across all the patients, in order to largely avoid noises. By heuristics, we set the frequency threshold for selecting phenotypes as 20. Similarly, for genetic features, we only kept the ones that appear more than 10 times across all the patients.

Associations between the patients and the features were captured to build the feature graph. Specifically, we made a link between phenotypic feature and genetic feature if both of them are associated with the same patient. As a result, the feature graph can be defined as \(G=(N,L)\), where \(N\) represents the set of phenotypic/genetic nodes and \(L\)
represents the set of links/edges between these nodes. We viewed graph \(G\) as an undirected graph that bridges mutations and phenotypes based on information collected from heterogeneous cancer patients in clinical practice.

\subsection{GNN Models}

\subsubsection{AGNN}

AGNN is a GNN model where the relations between the nodes are denoted
by the differentiable attention mechanism. The whole model is end-to-end
trainable, as all the functions in AGNN are parameterized by neural
networks. Given any two phenotypic/genetic feature representations \(f_i\), \(f_j\), and their edge representation \({e}_{i, j}\), the message aggregation is done as follows:

\begin{equation}
    \textbf{m}_i^k = \sum_{f_j \in \mathcal{N}_i} \textbf{m}_{j, i}^k
                   = \sum_{f_j \in \mathcal{N}_i} M(\textbf{h}_j^{k - 1}, \textbf{e}_{i, j}^{j - 1}),
\end{equation}

where \(M()\) is a message function, \(\textbf{m}_i^k\) is a received message, \(\mathcal{N}_i\) represents all the neighbor nodes for \(f_i\), and \(k\) indicates the number of steps. Given a state update function \(T()\), the node representation is updated as follows:

\begin{equation}
    \textbf{h}_i^{k} = T(\textbf{h}_i^{k - 1}, \textbf{m}_i^k),
\end{equation}

where \(h_i^0 = f_i\). After \(k\) iterations of
aggregation, \(\textbf{h}_i^k\) captures the relations within the \(k\)-hop
neighborhood of node \(f_i\).

\subsubsection{ChebNet}

ChebNet is a spectral-based graph learning algorithm. It is also a highly
efficient version of GCNN~\cite{gcnn}. Its convolutions are defined in a similar
fashion as they are usually defined on images, where a pixel value
is multiplied by the value of the kernel. In ChebNet, spectral convolutions
are defined as the multiplication of a signal by a kernel. Kernel is
comprised of Chebyshev polynomials. The polynomials are derived from
the diagonal matrix of Laplacian eigenvalues. The kernel can be represented
as the truncated expansion of order \(K - 1\) as follows:

\begin{equation}
    g_\theta(\Lambda) = \sum_{k=0}^{K - 1} \theta_k T_k(\tilde{\Lambda}),
\end{equation}

where \(g_\theta\) is a kernel applied to \(\Lambda\), the diagonal matrix of
Laplacian eigenvalues, and \(\theta \in \mathbb{R}^K\) is the vector of Chebyshev coefficients.
\(\tilde{\Lambda}\) represents the diagonal matrix of the scaled Laplacian eigenvalues.
\(K\) and \(k\) represent the largest and the smallest order neighborhoods respectively.
\(T_k(\tilde{\Lambda}) \in \mathbb{R}^n \times \mathbb{R}^n\) is the Chebyshev polynomial
of order k. Finally, the kernel is the sum of Chebyshev polynomials from \(k\) to \(K - 1\).

Hence, the ChebNet convolution is:

\begin{equation}
    g_\theta(\Lambda) \circ x = \sum_{k=0}^{K - 1} \theta_k T_k(\tilde{\Lambda}) x,
\end{equation}

where x is the signal. In this equation, \(\theta\)'s are the parameters to be trained.
It should also be noted that the representation of \(\tilde{\Lambda}\) maps the eigenvalues
in the interval \([-1 , 1]\). Such representation of the convolution greatly decreases
the number of computations to be performed.

\subsubsection{GCN}

GCN can be viewed as a simplified version of ChebNet. Its convolution is defined as follows:

\begin{equation}
    g_{\theta^{\prime}}(\Lambda) \star x = \sum_{k=0}^{K - 1} \theta_k^{\prime} T_k(\tilde{L}) x,
\end{equation}

where \(g_0\) is a kernel which is multiplied by the graph signal \(x\).
\(\theta_k^{\prime}\) is a vector of Chebyshev coefficients.
\(K\) is the \(K\)th order neighborhood and \(k\) is the closest order neighborhood.
\(T\) is the Chebyshev polynomial applied to \(\tilde{L}\).

ChebNet and GCN are similar. The biggest difference is in the choice for the value of K.
In the GCN, the layer wise convolution is limited to K = 1. This helps avoid overfitting
on the graph's local neighborhood.

\subsubsection{GAT}

The Graph Attention Network or GAT is a non-spectral learning method which
utilizes the spatial information of the node directly for learning. This is
in contrast to the spectral approach of the GCN which mirrors the same basics
as the CNN. On the other hands, GAT, similar to GCN, uses node self-attention
features and neighbor features for training the model. It also uses multi-head
attention (similar to models such as BERT~\cite{bert}).

The first step performed by the Graph Attention Layer is to apply a linear
transformation - Weighted matrix \(\textbf{W}\) to the feature vectors of the
nodes.

GAT uses attention coefficients for getting the relative importance
of neighboring features with respect to each other. For neighboring nodes
\(f_i\) and \(f_j\), the attention coefficient is calculated as follows:

\begin{equation}
    e_{i, j} = a(\textbf{W}\overrightarrow{h_i}, \textbf{W}\overrightarrow{h_j}),
\end{equation}

where \(a = \mathbb{R}^{F^{\prime}} \times \mathbb{R}^{F^{\prime}} \rightarrow \mathbb{R}\)
is shared attention mechanism, \(h_i, h_j\) are input features, and \(\textbf{W}\) is
a weighted matrix.

For making the attention coefficients easily comparable across different
feature nodes, GAT normalizes them across all choices of \(f_j\) by applying the
softmax function. Finally, the aggregation is done as follows:

\begin{equation}
    \overrightarrow{h_i^{\prime}} = \sigma \bigg(\frac{1}{K}\sum_{k = 1}^{K} \sum_{f_j \in \mathcal{N}_i}\alpha_{ij}^k\textbf{W}^k\overrightarrow{h}_j )\bigg)
\end{equation}

\subsubsection{GIN}

GIN is another graph neural network that updates the phenotypic/genetic node representations using the
multi-layer perceptron according to the following equation:

\begin{equation}
    h_i^{(k)} = \textnormal{MLP}^k\bigg(\big(1 + \epsilon^{(k)}\big) \cdot h_i^{k + 1} + \sum_{u \in \mathcal{N}(i)} h_j^{(k - 1)}\bigg),
\end{equation}

where \(h_i^{(k)}\) is the feature vector of node \(f_i\) at the \(k\)-th iteration/layer,
\(\mathcal{N}(i)\) is the set of nodes adjacent to \(f_i\), and MLP is a multi-layer
perceptron. Aggregation is done as follows:

\begin{equation}
    h_G = \textnormal{CONCAT}\bigg(\textnormal{READOUT}\bigg(\bigg\{h_i^{(k)}i \in G\bigg\}\bigg) \mid k = 0, 1, \dots, K),
\end{equation}

where CONCAT aggregates the embeddings of feature nodes. In this study, we adopted mean aggregation operation for GIN to learn phenotypic and genetic feature representations .

\subsubsection{GraphSAGE}

GraphSAGE is a framework for inductive representation learning.
It is suited for generating low-dimensional embeddings for nodes
and for graphs that grow in size over time. Since GraphSAGE is inductive
by nature, generating vector representation for new nodes is a lot faster
compared to transductive techniques such as DeepWalk~\cite{deepwalk}.
In this study, each phenotypic/genetic feature node is represented by the aggregation of its neighborhood. Specifically, we used GraphSAGE with the mean aggregator. Given any feature node \(f_i\), the equation for the mean aggregation is defined as:

\begin{equation}
    h_k^v \leftarrow \sigma(W \cdot \textnormal{MEAN}(\{h_i^{k - 1}\} \cup \{h_i^{k - 1}, \forall i \in \mathcal{N}(i)\})
\end{equation}

\subsubsection{SGC}

SGC is a simplified version of GCN where nonlinearities are successively
removed and weight matrices between consecute layers are collapsed. It is comprised of a graph-based
preprocessing step followed by a standard multi-class logistic regression, which is suitable for our task on multi-class cancer classification.

\subsubsection{TAGCN}

TAGCN is a graph neural network where each neuron in the graph convolutional layer
is connected only to a local region (local phenotypic/genetic feature nodes and links between them) in the vertex domain
of the input data volume, which is adaptive to the graph topology. It uses
a set of fixed-size learnable filters to perform convolutions on graphs.
In this study, the topologies of these filters are adaptive to the topology of the entire feature graph when they scan the graph to perform convolution.

\section{Evaluation}

We evaluated the results using metrics including accuracy, precision, recall,
and F1 score. For further performance assessment, ROC curve was also generated
for each model. Equations below describe these metrics:

\begin{align}
    & Accuracy  = \frac{TP + TN}{TP + TN + FP + FN}\\
    & Precision = \frac{TP}{TP + FN}\\
    & Recall    = \frac{TP}{TP + FN}\\
    & F1        = \frac{2 \times Precision \times Recall}{Precision + Recall}
\end{align}

\section{Results}
Table 1 shows the metrics for the GNN models. Additionally,
it shows the metrics for the baseline models including decision tree, gradient boosting,
multi-layer perceptron, naive bayes, and random forest. Our goal was to also compare the
results with those of more traditional machine learning techniques.

\begin{table}[h!]
\centering
    \caption{Evaluation results for the eight GNN models and five baseline models.}
    \begin{tabular}{ccccc}
        Model                   &       Accuracy &      Precision &         Recall &      F1 Score\\
        \hline
        AGNN                    &         0.877  &         0.890  &         0.889  &         0.885\\
        ChebNet                 & \textbf{0.901} &         0.902  & \textbf{0.920} &         0.907\\
        GAT                     &         0.753  &         0.726  &         0.684  &         0.697\\
        GCN                     &         0.802  &         0.855  &         0.787  &         0.804\\
        GIN                     &         0.827  &         0.821  &         0.809  &         0.805\\
        GraphSAGE               & \textbf{0.901} & \textbf{0.926} &         0.910  & \textbf{0.915}\\
        SGC                     &         0.827  &         0.836  &         0.832  &         0.824\\
        TAGCN                   & \textbf{0.901} &         0.907  &         0.910  &         0.904\\
        \hline
        Decision Tree           &         0.700  &         0.737  &         0.735  &         0.722\\
        Gradient Boosting       &         0.877  &         0.900  &         0.868  &         0.871\\
        Multi-layer Perceptron  &         0.864  & \textbf{0.902} &         0.851  &         0.870\\
        Naive Bayes             &         0.531  &         0.533  &         0.444  &         0.425\\
        Random Forest           & \textbf{0.888} &         0.897  & \textbf{0.895} & \textbf{0.888}\\
        \hline
    \end{tabular}
\end{table}

ChebNet, GraphSage, and TAGCN showed the best performance accuracy-wise,
all having roughly the same accuracy value of 0.901. AGNN came next with
the accuracy value of 0.877. GIN and SGC both had the approximate accuracy
of 0.827. GCN came next with the accuracy value of 0.827. The worst accuracy
was demonstrated by GAT, with the score of 0.753. As compared to accuracy
values of the baseline models, three GNN models outperformed random forest
classifier, which had the highest accuracy. Additionally, the mean accuracy
score for GNNs was approximately 0.849 while the mean accuracy value for the
baseline models was roughly 0.772.

GraphSAGE had the highest precision value of 0.926. TAGCN and ChebNet came
next with the values of 0.907 and 902 respectively. AGNN and GCN had the
precision values of 0.890 and 0.855 respectively. SGC had the precision value
of 0.836 and GIN's precision was 0.821. GAT had the precision value 0.726,
which was the lowest among the GNNs. Across all baseline models, multi-layer
perceptron had the highest precision value of 0.902 and gradient boosting
came close with the value of 0.900. GraphSAGE and TAGCN outperformed the
best baseline and ChebNet had the same approximate value. The mean precision
for GNNs was 0.858 and the mean precision for the baseline models was 0.794.

As for the recall, ChebNet showed the best performance among the GNNs, with
the value of 0.920. GraphSAGE and TAGCN both had the same recall value of
0.910. AGNN came next with the value of 0.889. SGC had the value of 0.832.
GIN and GCN had the recall values of 0.809 and 787 respectively. GAT had
the value of 0.684. The highest value for the baseline models was 0.895,
with the random forest classifier. ChebNet, GraphSAGE, and TAGCN had the
higher recall values. The mean recall values for GNN and baseline models
were 0.843 and 0.759 respectively.

Across all GNNs, GraphSAGE had the highest F1 score, roughly equal to 0.915.
ChebNet and TAGCN came next with the values of 0.907 and 0.904 respectively.
AGNN had the F1 score of 0.855. SGC and GIN had the scores of 0.824 and 0.805
respectively. GCN had the score of 0.804. GAT had the worst F1 score of 0.697.
Random forest classifier had the highest F1 score among all baseline models,
with the value of 0.888. Among GNNs, GraphSAGE, ChebNet, and TAGCN had higher
scores. The mean F1 score for the GNN models was 0.843 and the mean F1 score
for the baseline models was 0.755.

Overall, GNN models, on average, showed better performance in all metrics
than the baseline models. The mean values for all four metrics (accuracy,
precision, recall and F1 score) were higher for GNN models. Among the GNN
models, ChebNet, GraphSage, and TAGCN had the best overall performance and
GAT showed the worst overall performance.

\begin{table}[h!]
\centering
    \label{evaluation.diseases.table}
    \caption{Disease-specific evaluation of models based on F1 scores.}
    \begin{tabular}{cccccccc}
        Model                  &          Lung &      Prostate &        Breast &       Ovarian &      Pancreas &  Colon/Rectum &       Liver \\
        \hline
        AGNN                   &         0.89  & \textbf{0.95} &         0.83  & \textbf{0.87} &         0.93  &         0.88  &         0.85\\
        ChebNet                &         0.89  &         0.90  & \textbf{0.93} &         0.86  & \textbf{1.00} &         0.92  &         0.85\\
        GAT                    &         0.29  &         0.70  &         0.80  &         0.80  &         0.77  &         0.67  &         0.86\\
        GCN                    &         0.67  &         0.90  &         0.77  &         0.81  & \textbf{1.00} &         0.62  &         0.86\\
        GIN                    &         0.67  &         0.76  &         0.85  &         0.86  &         0.75  &         0.92  &         0.83\\
        GraphSAGE              & \textbf{1.00} &         0.90  & \textbf{0.93} & \textbf{0.87} &         0.93  &         0.92  &         0.85\\
        SGC                    &         0.67  &         0.90  &         0.86  &         0.81  &         0.93  &         0.71  & \textbf{0.89}\\
        TAGCN                  &         0.89  &         0.90  & \textbf{0.93} & \textbf{0.87} &         0.93  & \textbf{0.96} &         0.85\\
        \hline
        Decision Tree          &         0.80  &         0.67  &         0.71  &         0.63  &         0.86  &         0.67  &         0.75\\
        Gradient Boosting      & \textbf{0.89} &         0.86  & \textbf{0.89} &         0.86  &         0.77  & \textbf{0.91} & \textbf{0.92}\\
        Multi-layer Perceptron &         0.86  & \textbf{0.95} &         0.81  & \textbf{0.88} & \textbf{0.93} &         0.81  &         0.85\\
        Naive Bayes            &         0.00  &         0.35  &         0.65  &         0.67  &         0.62  &         0.44  &         0.25\\
        Random Forest          & \textbf{0.89} &         0.90  &         0.88  &         0.84  &         0.86  &         0.83  &         0.89\\
        \hline
    \end{tabular}
\end{table}

In the lung cancer classification task, GraphSAGE showed F1 score of roughly
1.00 and clearly outperformed any GNN or baseline model in this specific
disease classification task. AGNN, ChebNet, and TAGCN came next, all three
having the same the F1 score of 0.89. GCN, GIN, and SGC all had the F1 score
of 0.67. GAT showed the worst F1 score across the GNNs, approximately equal
to 0.29. Among the baseline models, gradient boosting and random forest showed
the best F1 scores, both having 0.89. They showed performance similar to AGNN,
ChebNet, and TAGCN, yet could not outperform GraphSAGE.

As for the prostate cancer, AGNN had the highest score across the GNNS, with
the value of 0.95. Multi-layer perceptron baseline had the same F1 score. ChebNet,
GCN, SGC, GraphSAGE, SGC, and TAGCN all had the same F1 scores of 0.90 which
was marginally worse than that of multi-layer perceptron baseline. GIN had the
lowest score among the GNNs, but performed well as compared with baseline models.
GAT had the worst performance among GNNs and second worst overall performance,
with the F1 score of 0.29.

In the breast cancer classification task, ChebNet, GraphSAGE, and TAGCN had
the same F1 score of 0.93. SGC and GIN had the F1 scores of 0.86 and 0.85
respectively. AGNN had the F1 score of 0.83 and GAT the F1 score of 0.80.
Among GNNs, GCN had the lowest F1 score, approximately equal to 0.77. The
highest score among baseline models was 0.89 with gradient boosting. ChebNet,
GraphSAGE, and TAGCN all had the higher F1 scores, but the rest of the models
showed a worse performance.

For the ovarian cancer, AGNN, GraphSAGE, and TAGCN all had the same F1 score,
equal to 0.87. ChebNet and GIN had marginally lower F1 scores, both equal to
0.86. GCN and SGC had the F1 scores of 0.81. GAT had the F1 score of 0.80.
Multi-layer perceptron was the baseline with the highest F1 score of 0.88.
It outperformed all GNN and baseline models.

In pancreas cancer classification, ChebNet and GCN had nearly perfect F1
scores, both approximately equal to 1.0. AGNN, GraphSAGE, SGC, and TAGCN
came next with the same score of 0.93. GAT and GIN had the F1 scores of
0.77 and 0.75 respectively. Among the baseline models, multi-layer perceptron
showed the highest score of 0.93, which was lower than all ChebNet and GCN,
but on par with AGNN, GraphSAGE, SGC, and TAGCN and higher that both GAT
and GCN.

As for colon/rectum cancer classification, TAGCN had the highest F1 score
among the GNNs, equal to 0.96. ChebNet, GIN, and GraphSAGE had the same F1
score of 0.92. AGNN had the F1 score of 0.88. SGC, GCN, and GAT had the scores
of 0.71, 0.67, and 0.62 respectively. Gradient boosting had the highest F1
score across all baselines, with the value of 0.91.

In the liver cancer classification task, SGC had the highest F1 score, equal
to 0.89. GAT and GCN came next with the same F1 score of 0.86. AGNN, ChebNet,
GraphSAGE, and TAGCN all had marginally lower F1 score of 0.85. GIN had the
score of 0.83. Interestingly, gradient boosting outperformed all GNN as well
as baseline models with the F1 score of 0.92.

To recapitulate, out of the seven classes, in four (lung cancer, breast cancer,
pancreas cancer, and colon/rectum cancer), the highest F1 scores came from GNNs.
In one of the classes (prostate cancer), the highest overall score was present
among both GNNs and baseline models had the same F1 score. In three of the classes
(ovarian cancer and liver cancer), highest F1 scores came from baseline models.

\section{Conclusions}

We harvested a cancer dataset comprising of 7 cancer disease classes and
applied 8 GNNs - AGNN, ChebNet, GAT, GCN, GIN, GraphSAGE, SGC, and TAGCN -
for disease classification purposes. We assessed the performance of the proposed
models using various metrics as well as compared them to the baseline models
including decision tree, gradient boosting, multi-layer perceptron, naive bayes,
and random forest. In this study, we discovered the potential usage of applying deep learning algorithms over joint phenotypic and genetic features for a better cancer type characterization. Moreover, we observed that ChebNet, GraphSage, and TAGCN outperformed other GNNs and baseline models, showing strong potential to be used for cancer type classification tasks.

\section*{Acknowledgements}
This work has been supported by the Gerstner Family Foundation, Center for Individualized Medicine (CIM) of Mayo Clinic.



\end{document}